%% file: main.tex
\documentclass{article}

 \usepackage[preprint]{style_files/neurips_2025}

\usepackage[utf8]{inputenc} 
\usepackage[T1]{fontenc}    
\usepackage{times}
\usepackage{latexsym}
\usepackage{microtype}
\usepackage{inconsolata}
\usepackage{graphicx}
\usepackage[most]{tcolorbox}
\usepackage{hyperref}       
\usepackage{url}            
\usepackage{booktabs}       
\usepackage{amsfonts}       
\usepackage{amsmath}
\usepackage{nicefrac}       
\usepackage{xcolor}         
\usepackage{algorithm}
\usepackage{algpseudocode}
\usepackage{tikz}
\usepackage{subfigure}
\usepackage{caption}
\usepackage{subcaption}
\usepackage{graphicx}
\usepackage{listings}
\usepackage{pgf-pie}
\usepackage{wheelchart}
\usetikzlibrary{positioning, backgrounds, graphs,arrows.meta, fit}
\usepackage{pgfplots}
\usepackage{adjustbox}
\pgfplotsset{compat=1.18}

\lstdefinelanguage{json}{
    basicstyle=\ttfamily\small,
    numbers=left,
    numberstyle=\tiny,
    stepnumber=1,
    numbersep=8pt,
    showstringspaces=false,
    breaklines=true,
    frame=lines,
    backgroundcolor=\color{gray!10},
    literate=
     *{0}{{{\color{blue}0}}}{1}
      {1}{{{\color{blue}1}}}{1}
      {2}{{{\color{blue}2}}}{1}
      {3}{{{\color{blue}3}}}{1}
      {4}{{{\color{blue}4}}}{1}
      {5}{{{\color{blue}5}}}{1}
      {6}{{{\color{blue}6}}}{1}
      {7}{{{\color{blue}7}}}{1}
      {8}{{{\color{blue}8}}}{1}
      {9}{{{\color{blue}9}}}{1}
      {:}{{{\color{red}:}}}{1},
}
\tikzset{
  barlabels/.style={font=\footnotesize\sffamily},
  declare function={
    barheight=5pt;
  }
}

\title{Mathematical Derivation Graphs: A Relation Extraction Task in STEM Manuscripts}

\author{%
  Vishesh Prasad 
    \\
  Electrical and Computer Engineering\\
  University of Illinois at Urbana-Champaign\\
  Urbana, IL 61801 \\
  \texttt{vprasad3@illinois.edu} \\
  \And
  Brian Kim \\
  Electrical and Computer Engineering\\
  University of Illinois at Urbana-Champaign\\
  Urbana, IL 61801 \\
  \texttt{brianhk2@illinois.edu} \\
  \AND
  Nickvash Kani \\
  Electrical and Computer Engineering\\
  University of Illinois at Urbana-Champaign\\
  Urbana, IL 61801 \\
  \texttt{kani@illinois.edu} \\
}

\begin{document}

\maketitle

\input{arxiv_preprint_2/sections/abstract}
\input{arxiv_preprint_2/sections/introduction}
\input{arxiv_preprint_2/sections/related_works}
\input{arxiv_preprint_2/sections/dataset}
\input{arxiv_preprint_2/sections/methods}

\input{arxiv_preprint_2/sections/results}
\input{arxiv_preprint_2/sections/conclusion}
\input{arxiv_preprint_2/sections/limitations}

\bibliographystyle{style_files/plainnat}
\bibliography{references}


\appendix

\input{arxiv_preprint_2/appendix/dataset}
\input{arxiv_preprint_2/appendix/algos}

\input{arxiv_preprint_2/appendix/large_language_models}
\input{arxiv_preprint_2/appendix/further_analysis}


\end{document}

%% file: arxiv_preprint_2/sections/abstract.tex
\begin{abstract} \label{sec:abstract}
Recent advances in natural language processing (NLP), particularly with the emergence of large language models (LLMs), have significantly enhanced the field of textual analysis. However, while these developments have yielded substantial progress in analyzing natural language text, applying analysis to mathematical equations and their relationships within texts has produced mixed results. This paper takes the initial steps in expanding the problem of relation extraction towards understanding the dependency relationships between mathematical expressions in STEM articles. The authors construct the Mathematical Derivation Graphs Dataset (MDGD), sourced from a random sampling of the arXiv corpus, containing an analysis of $107$ published STEM manuscripts with over $2000$ manually labeled inter-equation dependency relationships, resulting in a new object referred to as a derivation graph that summarizes the mathematical content of the manuscript. The authors exhaustively evaluate analytical and machine learning (ML) based models to assess their capability to identify and extract the derivation relationships for each article and compare the results with the ground truth. The authors show that the best tested LLMs achieve $F_1$ scores of $\sim45\%-52\%$, and attempt to improve their performance by combining them with analytic algorithms and other methods.
\end{abstract}

%% file: arxiv_preprint_2/sections/introduction.tex
\section{Introduction} \label{sec:introduction}
\begin{figure*} [htbp]
    \centering
    \fbox{\adjustbox{padding=5pt}{\includegraphics[width=1\linewidth]{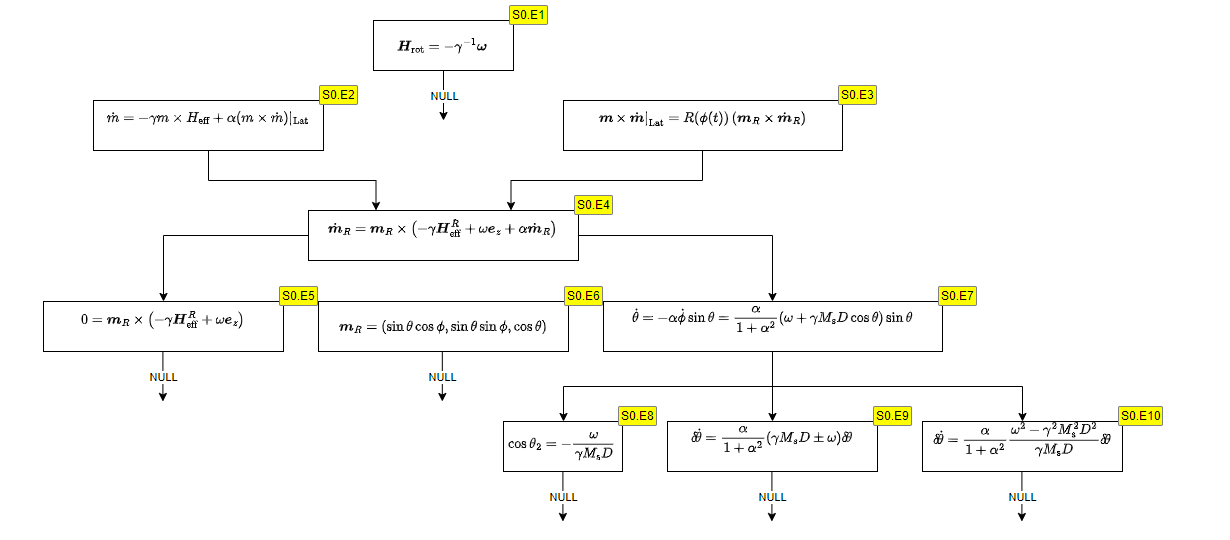}}}
    \caption{Ground truth derivation graph from article 0907.2648 \citep{bretzel2009barnett}. Each node represents a key equation in the article, and edges represent derivation relationships.}
    \label{fig:derivation_graph}
\end{figure*}
A known problem in natural language processing (NLP) is the task of information extraction (IE) \citep{sarawagi2008information}, which attempts to extract meaningful information from unstructured texts about entities. A subproblem of IE is relation extraction (RE), which  extracts structured relations and connections between entities in texts specific to a domain \citep{10.5555/1071958, 10.5555/1072064, 10.5555/1072017, 10.5555/1072399, chinchor-1998-overview}. RE identifies and extracts semantic relationships between entities in text, allowing for knowledge-graph construction, search, and other downstream applications \citep{qu2020fewshotrelationextractionbayesian, Chen_2022, ji2021survey, yu2020relationship}. RE methods can range from rule-based and pattern approaches \citep{riloff1993automatically, hobbs1993fastus, muslea1999extraction, ciravegna2001adaptive}, up to modern transformer-based large language models \citep{wan2023gptreincontextlearningrelation, wadhwa-etal-2023-revisiting}.

In this paper, the authors present a new formulation of relation extraction in relation to the mathematical content of STEM manuscripts. In STEM multi-modal (natural language text + symbolic math) manuscripts, uncovering the relationships between mathematical equations can help readers and systems trace logical dependencies and intermediate steps, allowing for equation-centered search \citep{10.1007/s10791-017-9296-8}, proof understanding \citep{wang2023legoproverneuraltheoremproving}, formula retrieval \citep{9671942}, and other use cases.

\textit{In this manuscript, the authors propose a new problem: extracting derivation graphs, a directed acyclic dependency graph  representing lengthy, mathematical derivations whose nodes represent equations within the manuscript and edges represent a derivative relationship where one equation was used to formulate the new equation.} Representing mathematical manuscripts as derivation graphs and their construction as an RE problem makes it possible to apply structured analytical and machine learning methods to find derivative relationships in manuscripts.

The authors' contributions are twofold: \textit{(i) the authors have constructed a new hand-annotated Mathematical Derivation Graphs Dataset (MDGD) of real-world arXiv documents containing lengthy, multi-step derivations and their corresponding derivation graphs, (ii) the authors explore multiple analytical and machine learning methods for automatically extracting these derivation graphs and measuring their efficacy}. Extraction techniques that the authors explore include rule-based baselines that leverage provided document information and modern large language models, which are prompted for the task. Performance is measured at the edge-level, using metrics such as precision, recall, and $F_1$ \citep{chinchor-1998-overview} for predicted dependency edges and relationships, and the authors analyze shortcomings qualitatively.

Modern large reasoning models increasingly depend on explicit step-by-step (chain-of-thought) decompositions to solve complex symbolic and mathematical problems \citep{chen2023programthoughtspromptingdisentangling}, yet existing training data rarely exposes the inter-equation dependencies needed to supervise such reasoning. Models typically learn these steps through distillations from larger models on datasets that often lack these step-by-step instructions \citep{zhu2024keypointdrivenmathematicalreasoningdistillation}. Derivation graphs explicitly provide this missing structure in an organized manner, showing how equations relate to one another and offering higher-quality supervision for training models to generate correct stepwise solutions. Our task addresses this critical gap by offering the fine-grained, derivation-level information needed to train and evaluate more capable mathematical AI systems.

Additionally, derivation graph extraction can be very helpful to STEM researchers by summarizing math-heavy texts and augmenting scientific research models \citep{Wang_2025, lala2023paperqaretrievalaugmentedgenerativeagent}.

The authors' comprehensive testing finds that both analytical and machine learning models (including LLMs) achieve mostly $\sim32\%-52\%$ $F_1$ scores for extracting derivation graphs from articles, revealing that the recent advances in NLP have not made significant inroads in comprehending mathematical texts compared to simpler analytic models. Motivated by error analyses, the authors propose and test targeted fixes, including edge-limit constraints, refined prompt engineering, hybrid analytic-LLM pipelines, few-shot prompting, and model fine-tuning. These refinements yield $F_1$ scores that slightly improve on the performance of base models, highlighting that more work needs to be done in this space with machine models for better mathematical relation extraction. The authors' findings highlight that while current approaches offer a solid foundation for extracting mathematical information and constructing a corresponding derivation graph, further research is necessary to improve accuracy and depth in this area.

\section{Problem Definition}
The authors define \textbf{key math equations} as those explicitly labeled with a reference number in the manuscript, as seen in Fig.~\ref{fig:edge_example}. Since manuscripts often contain hundreds of mathematical equations, many of which may be uninteresting (e.g., ``$k_b = 1.38 \cdot 10^{-23} J\cdot K^{-1}$''), the authors limit the formulation of the derivation graph to only contain the key math equations within a manuscript. Qualitative analysis has shown that these numbered equations help illustrate the manuscript's primary focus, making them a reliable indicator for inclusion in the derivation graph.

A derivation is defined as originating, developing, or being obtained from another object. In this context, a \textbf{derivation graph} is a directed acyclic graph $G = (V, E)$, where each node $v \in V$ represents a distinct mathematical equation or expression extracted from the article. Each directed edge $(v_i \rightarrow v_j) \in E$ indicates that equation $v_j$ is derived from or logically depends on equation $v_i$. Edges encode a notion of informational flow or computational dependency within the derivation graph. The graph is acyclic by design, as derivational dependencies do not form loops in logical progression. An example of an extracted derivation graph is shown in Fig.~\ref{fig:derivation_graph}, where nodes represent ``key math equations'' from the manuscript and directed edges between nodes, $\left(u, v\right)$, indicate that equation $v$ was derived from equation $u$ within the text.    

%% file: arxiv_preprint_2/sections/related_works.tex
\section{Related Works}

While the task of extracting equations and derivation relationships between these equations is a novel one, the task of extracting relationships between entities in natural language text is an ongoing topic of research \citep{DETROJA2023200244}. Relation extraction (RE), a subproblem of Information Extraction (IE), evolved from the Message Understanding Conferences (MUCs) \citep{10.5555/1071958, 10.5555/1072064, 10.5555/1072017, 10.5555/1072399, chinchor-1998-overview}, where groups were tasked with developing systems to extract specific relations. These MUCs led to many developments, such as Named Entity Recognition (NER) \citep{grishman-sundheim-1996-message}, Template Relations (TR) \citep{marsh-perzanowski-1998-muc}, and, importantly, the introduction of metrics on which RE tasks should be evaluated, such as Precision, Recall, and $F_ 1$-score \citep{chinchor-1998-overview}.

Traditional RE dealt with rule-based systems \citep{riloff1993automatically, hobbs1993fastus, muslea1999extraction, ciravegna2001adaptive}, where relations were chosen from pre-defined sets specific to a domain. These worked well in systems where textual patterns could be easily defined, but fell short in systems with undefined patterns and unseen corpora. In order to decrease the human effort needed to set up RE tasks, \cite{10.1145/1409360.1409378} introduced a new RE model named open information-extraction (OpenIE), where relations are automatically extracted without having pre-
defined relation sets or extraction patterns. The systems developed from this approach are able to extract patterns from labeled data or predefined patterns, and then use those patterns to extract relations from text. Recent developments have led to the emergence of deep learning \citep{jiang-etal-2016-relation, qin2017, stanovsky-etal-2018-supervised, zhan2019spanmodelopeninformation, zhong2021frustratinglyeasyapproachentity} paradigms and the application of large language models \citep{wan2023gptreincontextlearningrelation, wadhwa-etal-2023-revisiting} to the relation extraction problem. Although relation extraction has traditionally been limited to shorter texts, an extension of the RE problem involves document-level extractions \citep{10.1007/s10489-024-05985-y}. However, more research is needed on the relation extraction problem when relating to math-heavy manuscripts. Previous works have extended the relation extraction to enhance LLM's abilities with word problems \citep{zieword}, but this extension is limited to simple arithmetic reasoning in natural language word problems.

Current work on theorem-proving using LLMs \citep{wang2023legoproverneuraltheoremproving, zhou2024refactorlearningextracttheorems} populate their libraries by extracting theorems from technical literature. The primary focus of this work is the analysis of extracting reasoning steps from technical documents with regards to step-by-step proofs, allowing for the extension of derivation graphs to help with theorem proving, which is left to future work.

%% file: arxiv_preprint_2/sections/dataset.tex
\section[Dataset]{Dataset\texorpdfstring{\footnote{\url{https://github.com/visheshprasad/Mathematical_Derivation_Graphs_Dataset_MDGD}}}{}}
\label{sec:dataset}

To facilitate the authors' investigation into extracting mathematical relationships using derivation graphs, the authors developed a hand-labeled corpus named the Mathematical Derivation Graphs Dataset (MDGD), which was randomly sampled from the arXMLiv dataset \citep{arxiv, arXMLiv} of STEM articles under the CC-BY 4.0 Creative Commons Attribution license. The arXMLiv dataset is a valuable resource that provides an extensive collection of articles from the arXiv preprint archive, converted into HTML and MathML \citep{w3cMathML3} format. This HTML and MathML conversion allows for easier parsing and processing of textual content and mathematical expressions, making it a suitable choice for the task. The dataset encompasses various scientific disciplines, enabling us to capture diverse mathematical derivations across multiple domains. 

To ensure a comprehensive and manageable dataset, selective inclusion criteria were employed in the construction of the corpus. The authors randomly sampled articles from the arXMLiv dataset, ensuring that the selected samples were met the criteria of (1) the number of equations being between $5$ and $20$ equations to balance the complexity of the articles for hand labeling, while ensuring that they were mathematically rich enough to provide meaningful insights during the extraction process, and (2) formatting to ensure that the articles could be parsed correctly. To ensure that the derived graphs are generalizable across different disciplines, articles were randomly sampled from various categories, including Physics, Mathematics, Computer Science, and Biology. A breakdown of the domain-wise distribution of the dataset is found in Figure~\ref{fig:dataset_categories}. The authors cover the disciplines most likely to benefit from derivation graphs, and since other fields may not benefit as much, they constitute a proportionally smaller subset of the dataset. With this dataset construction, the authors ended up with a total of $107$ articles, totaling over $2000$ equations. 

\begin{figure}[htbp]
    \centering
    \begin{tikzpicture}[
      y=0.3cm,
      x=0.06cm,
    ]
    \foreach [count=\i from 0] \p/\t in
                      {28/Condensed Matter,
                      17/Astrophysics,
                      17/Physics,
                      10/High Energy Physics - Phenomenology,
                      8/Quantum Physics,
                      7/Computer Science,
                      5/Mathematics,
                      3/General Relativity and Quantum Cosmology,
                      3/Nuclear Theory,
                      2/High Energy Physics - Lattice,
                      2/Nonlinear Sciences,
                      1/Electrical Engineering and Systems Science,
                      1/High Energy Physics - Theory,
                      1/Mathematical Physics,
                      1/Nuclear Experiment,
                      1/Quantitative Biology
                       }
      {
       \node [anchor=base east,
              barlabels,
              name=i-\i] at (0,-\i) {\t};
       \fill [blue!40] (i-\i.base east) rectangle ++(\p,barheight)  ++(0,-barheight)
              node[barlabels, 
                   black,
                   anchor=base west] {\p};
      }
    
    \end{tikzpicture}
    \caption{Primary category count breakdown for all articles in the MDGD \citep{math_derivation_graphs} based on the arXiv category taxonomy (n=107).}
    \label{fig:dataset_categories}
\end{figure}

Each article's entry in the dataset comprises four fields: ``Article ID'', ``Equation ID'', ``Adjacency List'', ``Equation Number''. Details on the specifics of what each field can be found in Appendix~\ref {apx:data}. A specific JSON example of a dataset entry is shown in Fig.~\ref{fig:json_example}. For each article, the authors hand-annotated the article and filled in each individual entry in the dataset. To ensure the validity and consistency of the corpus, a random subset of articles was processed by all annotators. Additionally, the annotators authenticated each other's labeling of every corpus entry multiple times during the dataset creation process.

The corpus contains $107$ manually annotated articles. Two primary annotators performed all labeling, while a third served as a mediator and adjudicated disagreements by majority vote. To check consistency, a random subset of articles were independently labeled by all annotators, and every corpus entry was cross-checked by the annotators during the labeling process. Out of 107 articles, the annotators disagreed on $12$ $(12/107 \approx 11.2\%)$, yielding a raw agreement rate of $95/107 \approx 88.8\%$. No automated annotation tool was used. When dealing with human annotators, it is inevitable that some subjectivity and biases will be present. To minimize these, the annotators held multiple discussions to ensure consistency with the annotation goals, guided by the principle of minimizing the derivation graph's complexity while preserving the semantic structure of the underlying manuscript.

Ultimately, the constructed Mathematical Derivation Graphs Dataset (MDGD) forms the foundation of the study, enabling rigorous evaluation of the effectiveness of various algorithmic approaches for extracting mathematical relationships. The selection of these $107$ articles, coupled with the focus on diversity and manageable complexity, allows for a balance between comprehensiveness and feasibility, laying the groundwork for exploring more advanced techniques in future work.

%% file: arxiv_preprint_2/sections/methods.tex
\section{Methods} \label{sec:models}

In this section, the authors evaluate different methods on the relation extraction task of extracting the derivation relations for articles from the MDGD dataset presented in \S~\ref{sec:dataset}. The authors' study performs initial experiments for this task using various methods, including analytic, machine-learning-based, and LLM models. The authors also perform an initial analysis on the methods used and test potential fixes as a first step toward solving this task. Further analysis of the algorithms are presented in \S~\ref{sec:results}.

\subsection{Analytical Methods}
The authors tested multiple analytical methods to provide a baseline for low-cost methods in this relation extraction task. 

The \textbf{Brute Force} method leverages a simple but effective insight: the most obvious connections between equations are often directly stated in the surrounding text. This method takes advantage of this by scanning the sentences immediately before and after each equation for explicit mentions of other equations, as shown in Fig.~\ref{fig:edge_example}. These explicit references often indicate a direct derivation relationship and provide an easy starting point for edge construction in the graph. Implementing the brute force algorithm (details found in Appendix~\ref{apx:dga}) reveals a consistent linguistic structure surrounding many of these edges. Sentences that describe derivations often conclude with highly predictable phrases such as ``yields,'' ``becomes,'' ``proves,'' ``get,'' ``takes the form,'' ``find that,'' or ``write as.'' These markers are strong indicators of the logical progression from one equation to the next and can be exploited to extract derivation links more reliably. Further improvements may arise from a more fine-grained analysis of the textual context surrounding these equations, particularly in determining edge directionality when multiple equations are mentioned together.

\begin{figure}[htbp]
    \centering
    \fbox{
        \begin{minipage}{0.95\linewidth}
            where \(\beta\) is a constant velocity anisotropy parameter. Combining the mass density profiles in Eq. (4),
            \begin{equation}
                M(r) = \frac{2}{\sqrt{\pi}\lambda(\alpha)}\left(\frac{r}{R_E}\right)^{3-\alpha}M_E \tag{8}
            \end{equation}
        \end{minipage}
    }
    \caption{A brute force derivative edge from Eq.~(4) to Eq.~(8) in article 1701.00003 \citep{zhu2017local}. Eq.~(8) is implied as being key since it is explicitly labeled with a reference number. The definite edge shown in the text before the equation represents a derivation relationship from Eq.~(4), pointing to Eq.~(8).}
    \label{fig:edge_example}
\end{figure}

The \textbf{Token Similarity} method builds on the observation that equations derived from one another often reuse content. The token similarity algorithm constructs edges in the derivation graph by comparing the degree of token overlap between two equations. With the author's implementation (details found in Appendix~\ref{apx:dga}), tokenization for any equation is done at the character level. Rather than relying on token order, it checks whether the tokens from one equation appear in the other. When the overlap crosses a set threshold, an edge is inferred. This builds on the idea that equations derived from another often preserve parts of the original equation. An example can be seen in Fig.~\ref{fig:combined_vertical}, which illustrates an edge from Eq.~(3) to Eq.~(6) based on token overlap.

\begin{figure}[htbp]
    \centering
    \fbox{
        \begin{minipage}{0.95\linewidth}
            \begin{equation}
                T(A) = 4A \int_0^1 \frac{du}{\sqrt{e^{{A^2}(1-u^2)}-1}} \tag{3}
            \end{equation}
            In order to derive the small-amplitude expansion we change the integration variable in equation (3) to \(u = \cos \theta\) so that the period becomes
            \begin{equation}
                T(\rho) = 4 \int_0^{\pi / 2}\frac{d\theta}{\sqrt{F(\rho)\sin^2\theta}} \tag{6}
            \end{equation}
        \end{minipage}
    }
    \caption{Example of derivative relation found via token similarity between (3)and(6) in article 0907.3505, and confirmed by brute force edge\citep{fernández2010smallamplitudeapproximationdifferentialequation}.}
    \label{fig:combined_vertical}
\end{figure}

The \textbf{Naive Bayes} method utilizes the machine learning based Naive Bayes model \citep{mccallum1998comparison}. This model was a natural choice for this relation extraction task due to its efficiency and interpretability. The goal was to leverage the relationship between the text in the article and its mathematical equations to predict edges in a derivation graph. To train this model, the authors needed to construct labels and features from each article (implementation details can be found in Appendix~\ref{apx:dga}). To be input into the model, each feature is converted into a bag-of-words count vector, and a MultinomialNB model is trained on these count vectors. The authors trained the model using a percentage of the MDGD data. Interestingly, reducing the training set size had a marginal effect on accuracy and other metrics, suggesting that the extracted features carry strong predictive power. While the conditional independence assumption makes the model elegant and fast, it does not always hold in practice. Mathematical or scientific writing often carries context that depends on surrounding words. As such, there is room for improvement through better feature engineering, such as experimenting with broader or narrower text windows, filtering for more relevant content, or incorporating structural cues from the article layout.

\subsection{Large Language Models}
\textbf{Large Language Models} (LLMs) were used primarily to benchmark the previously discussed analytical and machine learning methods against state-of-the-art language and transformer models at the forefront of current research. The authors ran experiments on extracting derivation graphs from the MDGD dataset with six popular models: Google's Gemini 2.5 flash \citep{comanici2025gemini25pushingfrontier}, OpenAI's GPT-5 \citep{openai2025gpt5}, Meta's Llama-3.2-3B-Instruct \citep{touvron2023llama2openfoundation}, Mistral-Nemo-Instruct-2407 by Mistral AI and NVIDIA \citep{mistral2024}, Qwen2.5-Coder-32B-Instruct from Alibaba and Qwen \citep{hui2024qwen25codertechnicalreport}, and Microsoft's Phi-3.5-mini-instruct \citep{haider2024phi3safetyposttrainingaligning}. These models were evaluated without fine-tuning and used as-is, with their original number of parameters, through Google's, OpenAI's, and Hugging Face's APIs. The initial experimentation with these LLM models employed zero-shot evaluation to establish an initial performance benchmark. Models that performed well advanced to few-shot and combination experiments seen in \S~\ref{subsec:dg_i_pf}. This two-stage testing protocol balances comprehensive testing coverage of this extraction task with the additional computation cost needed to evaluate more resource-intensive experiments.

An important consideration when using LLMs is the design and engineering of prompts, which significantly impact consistency. After trying several formats (further details in Appendix~\ref{apx:llm}) on all LLM models, the best-performing version is found in Fig.~\ref{fig:prompt-box}. This prompt yielded the most parsable output ($\sim93\%$ of outputs were formatted correctly), which was important since the authors needed structured, machine-readable results. Several alternative phrasings were tested (see Appendix~\ref{apx:llm}) to evaluate the sensitivity of models to instruction design. Though some models performed better with different types of prompt phrases and article inputs, these alternate prompts did not consistently format the outputs correctly enough across any models to be considered feasible. The model's formatting failures were not included in the performance metric calculation. If a model returned an incorrectly formatted output, the authors simply disregarded it when calculating the metrics.

\begin{figure}[ht]
\centering
\begin{tcolorbox}[enhanced,
  attach boxed title to top center={yshift=-3mm,yshifttext=-1mm},
  colback=blue!5!white,
  colframe=blue!75!black,
  title=Best Found LLM Prompt]
``I have the following article that contains various mathematical equations: $\backslash$n''+\{total\_article\_text\}+``$\backslash$n From this article, I have extracted the list of equations, numbers as follows: $\backslash$n''+\{equation\_list\}+``$\backslash$n Analyze the context of the article to identify which equations are derived from each equation. Provide the output as a list and nothing else, with the format: w -> x, y, z;$\backslash$n x -> h, t;$\backslash$n ... If no equations are derived from a certain equation, return an empty list with the format: t ->;$\backslash$n''.
\end{tcolorbox}
\caption{Best performing prompt template for zero-shot extracting equation derivations from an article.}
\label{fig:prompt-box}
\end{figure}

%% file: arxiv_preprint_2/sections/results.tex
\begin{table*}[htbp]
  \caption{Initial Algorithm Performance Metrics for Equation Derivation Graph Extraction}
  \label{tab:derivation_correctness}
  \centering
  \begin{tabular}{p{6cm}|ccc}
    \toprule
    \hline
    \textbf{Algorithm} & \textbf{Precision} & \textbf{Recall} & $\mathbf{F_1}$ \\
    \midrule
    \hline
    Brute Force  & $\mathbf{49.4}\%$ & $\mathbf{49.4}\%$ & $\mathbf{49.4}\%$ \\
    Token String Similarity  & $40.5\%$ & $35.8\%$ & $38.0\%$ \\
    Naive Bayes  & $25.1\%$ & $47.6\%$ & $32.9\%$ \\ 
    \hline
    Zero-Shot - Gemini-2.5-flash  & $\mathbf{47.9}\%$ & $48.0\%$ & $47.9\%$ \\
    Zero-Shot - GPT-5  & $46.8\%$ & $\mathbf{58.5}\%$ & $\mathbf{52.0}\%$ \\
    Zero-Shot - Llama-3.2-3B-Instruct  & $16.0\%$ & $26.6\%$ & $20.0\%$ \\
    Zero-Shot - Mistral-Nemo-Instruct-2407  & $29.0\%$ & $22.0\%$ & $25.0\%$ \\
    Zero-Shot - Qwen2.5-Coder-32B-Instruct  & $43.8\%$ & $46.6\%$ &  $45.2\%$ \\
    Zero-Shot - Phi-3.5-mini-instruct  & $25.6\%$ & $57.9\%$ & $35.5\%$ \\
    \hline
    \bottomrule
\end{tabular}
\end{table*}

\section{Results} \label{sec:results}
A range of evaluation metrics, such as precision, recall, and $F_1$ score, were used to better understand how each derivation graph extraction algorithm performs, in terms of what it excels at and where it falls short. These metrics are chosen because they are the principal metrics by which RE tasks should be evaluated \citep{chinchor-1998-overview}.

\subsection{Overview}
As shown in Table~\ref{tab:derivation_correctness}, the Brute Force, Gemini, and GPT models outperform the others in terms of $F_1$ score. The numbers also reveal that the Brute Force approach achieves the highest precision, with approximately $50\%$ of the edges it predicts being indeed correct. However, the method still yields many false positives, likely due to identifying co-mentioned equations that do not have a derivation relationship. The authors also see that the GPT-5 LLM achieves the highest recall, successfully capturing a larger portion of true edges. This indicates its relative strength in identifying derivations that are implied but not explicitly stated in the surrounding text. In contrast, the Token String Similarity and Naive Bayes approaches perform comparatively worse. It can be hypothesized that the reason for this is that the former fails to consider textual context and relies solely on equation content, resulting in a high false positive rate. Despite functioning reasonably well given minimal training data, the Naive Bayes classifier lacks the capacity to incorporate contextual information, which limits its effectiveness. LLMs, for the most part, except for Gemini and GPT, performed terribly. Llama and Mistral achieved very low $F_1$ scores of $20\%$ and $25\%$, respectively, highlighting how models trained for language use cases have difficulty with mathematical aligned tasks. Additionally, these models struggled with consistent output formatting.

\subsection{Error Analysis} \label{subsec:dg_ir_ea}
To better understand the limitations of the evaluated methods, the authors qualitatively compared their output graphs with the ground truth (see Fig.~\ref{fig:output_comp}). This revealed several recurring issues, especially in LLM-based approaches:

\begin{enumerate}
    \item \textbf{Failure to Detect Explicit Edges:} LLMs frequently miss edges explicitly stated in the text, which are also edges reliably detected by the Brute Force algorithm. This could be due to: (a) \textbf{Contextual Overhead:} LLMs are optimized to interpret contextual clues and produce coherent responses. This strength can become a drawback when fine-grained, explicit relationships are lost amidst a broader context. (b) \textbf{Input Length Constraints:} Large input sizes may introduce noise, making it harder for the LLM to retain and focus on explicit derivation links, especially when the relevant information is dispersed.

    \item \textbf{Overgeneration of Implicit or Incorrect Edges:} LLMs tend to infer derivation edges between structurally similar equations, even when no such relationships exist. Such assumptions often result in excessive forward edges. Forward edges connect a parent node to its descendants, who can already be reached through the parent node's children. This issue can be attributed to: (a) \textbf{Pattern-Based Matching:} LLMs trained on massive corpora are inclined to connect patterns. Equations with similar structure or terms may be embedded closely, leading the LLM to assume they are semantically linked, even when they are not. (b) \textbf{Overreliance on Equation Embeddings:} The models may de-emphasize textual context, which is crucial for interpreting derivations, in favor of purely structural or symbolic cues from the equations, which inflates the number of falsely inferred edges.
\end{enumerate}

\begin{figure*}[htbp]
    \centering
\fbox{%
\adjustbox{padding=5pt}{%
\begin{tikzpicture}[
    >=Stealth,
    node distance=1cm and 1.5cm,
    every node/.style={draw, rectangle, minimum height=8mm, inner sep=2pt, font=\footnotesize},
]
  \usetikzlibrary{fit,positioning}

  \begin{scope}[local bounding box=GT]
    \node[fill=blue!20] (E1) {S2.E1};
    \node[fill=blue!20, below=of E1] (E2) {S2.E2};
    \node[fill=blue!20, right=of E2] (E4) {S2.E4};
    \node[fill=blue!20, above= of E4] (E3) {S2.E3};
    \node[fill=blue!20, right=of E3] (E5) {S3.E5};
    \node[fill=blue!20, right=of E4] (E6) {S3.E6};

    \draw[->] (E2) -- (E4);
    \draw[->] (E3) -- (E2);
    \draw[->] (E5) -- (E4);
    \draw[->] (E6) -- (E5);
  \end{scope}

  \node[draw=black, thick, fit=(GT), inner sep=8pt] (boxGT) {};
  \node[above=2pt of boxGT.north] {\small\bfseries Ground Truth Graph};

  \begin{scope}[xshift=7cm, local bounding box=COMP]
    \node[fill=green!30] (C1) {S2.E1};
    \node[fill=green!30, below=of C1] (C2) {S2.E2};
    \node[fill=green!30, right=of C2] (C4) {S2.E4};
    \node[fill=green!30, above=of C4] (C3) {S2.E3};
    \node[fill=green!30, right=of C3] (C5) {S3.E5};
    \node[fill=green!30, right=of C4] (C6) {S3.E6};

    \draw[->] (C1) -- (C2);
    \draw[->] (C2) -- (C4);
    \draw[->] (C3) -- (C2);
    \draw[->] (C4) -- (C5);
    \draw[->] (C4) -- (C6);
    \draw[->] (C6) -- (C5);
  \end{scope}

  \node[draw=black, thick, fit=(COMP), inner sep=8pt] (boxCOMP) {};
  \node[above=2pt of boxCOMP.north] {\small\bfseries Gemini-2.5-flash LLM Output};

  \node[align=center, font=\footnotesize] at (6,-3) {
    Precision = 0.50\quad Recall = 0.75\quad $F_1$ = 0.60
  };

\end{tikzpicture} }}
\caption{Derivation graph output comparison between the ground truth (left) and Gemini LLM (right) for article 1701.00226 \citep{Zhou_2018}. True positives (TP) $= 3$, false positives (FP) $= 3$, false negatives (FN) $= 1$. Precision $=$ TP / (TP + FP), Recall $=$ TP / (TP + FN), $F_1 = $ $2$ $\times$ Precision $\times$ Recall / (Precision $+$ Recall).}
\label{fig:output_comp}
\end{figure*}

\subsection{Proposed Fixes and Results} \label{subsec:dg_i_pf}
Based on the analysis in \S~\ref{subsec:dg_ir_ea}, the authors hypothesize and test specific improvements to improve performance results. The authors primarily focus on improving LLM performance, expecting it to yield better results. Results of the potential fixes can be found in Table~\ref{tab:fixes}.

\subsubsection{Combination Algorithms}
A proposed fix for the ``Failure to Detect Explicit Edges'' is the combination algorithm. Given the observation that the brute force algorithm can capture many of the explicit edges in the derivation graph, and LLMs may be better at identifying implicit edges, a potential solution is to combine both methods. The explicit edges would first be extracted using the brute force algorithm, then added to the LLM prompt, prompting the LLM model to add the rest of the implicit edges. Such a method would provide more structure to the LLM and could enhance its performance by leveraging its strengths. Combining the brute force algorithm with the LLMs increased the $F_1$ score for the Gemini-2.5-flash model compared to other potential fixes. This was mainly due to an increase in the recall, which indicates that the explicit edges provided are helping fill in the gaps for edges the LLM may be missing, and helping the LLM focus its context on finding more implicit edges. However, it is noted that this increase in $F_1$ score is accompanied by a slight decrease in precision compared to the original LLM performance. Such a drop would mean that the LLM is labeling slightly fewer edges as part of the derivation graph than it does regularly. This could also be due to certain edges picked up by the brute force algorithm that would not have been chosen by the LLM, resulting in a drop in precision.

\begin{table*}[htbp]
  \caption{Algorithm Performance Metrics for Potential Fixes}
  \label{tab:fixes}
  \centering
  \begin{tabular}{p{7cm}|ccc}
    \toprule
    \hline
    \textbf{Algorithm} & \textbf{Precision} & \textbf{Recall} & $\mathbf{F_1}$ \\
    \midrule
    \hline
    Brute Force + Gemini-2.5-flash  & $\mathbf{44.6}\%$ & $59.2\%$ & $50.8\%$ \\
    Brute Force + GPT-5  & $43.5\%$ & $\mathbf{62.4}\%$ & $\mathbf{51.3}\%$ \\
    \hline
    Just Equations - Gemini-2.5-flash & $43.7\%$ & $44.7\%$ & $44.2\%$ \\
    Two Sentences + Equation - Gemini-2.5-flash & $\mathbf{51.1}\%$ & $\mathbf{50.6}\%$ & $\mathbf{50.8}\%$ \\
    \hline
    Edge Limitation - Gemini-2.5-flash & $\mathbf{51.5}\%$ & $\mathbf{48.5}\%$ & $\mathbf{49.9}\%$ \\
    Best of Alternate Prompt - Gemini-2.5-flash & $42.4\%$ & $48.3\%$ & $45.2\%$c\\
    \hline
    Post-Processing - Brute Force + Gemini-2.5-flash & $48.7\%$ & $52.7\%$ & $50.6\%$ \\
    Post-Processing - Zero-Shot - GPT-5 & $\mathbf{48.9}\%$ & $\mathbf{55.9}\%$ & $\mathbf{52.2}\%$ \\
    \hline
    Few-Shot - Gemini-2.5-flash  & $\mathbf{49.8}\%$ & $49.9\%$ & $49.8\%$ \\
    Few-Shot - GPT-5  & $47.6\%$ & $\mathbf{57.2}\%$ & $\mathbf{52.0}\%$ \\
    \hline
    \bottomrule
\end{tabular}
\end{table*}

\subsubsection{Less Context}
An identified subproblem with ``Failure to Detect Explicit Edges'' is the contextual overhead issue. With the original prompt, the whole article's texts were given as additional context to the language model. Therefore, the authors experimented with decreasing context with two methods, the first being providing just the equations to the prompt, and the second one being only providing the sentence preceding and succeeding each equation, along with the equation itself. The difference in results between the above two experiments and the zero-shot experiment on the same model shows that some context is definitely needed (as evidenced by the drop in $F_1$ when feeding only the equations). When providing only one sentence on either side of the equation, the authors observe a slight increase in the $F_1$ score, which supports the hypothesis that the extra context is drowning out more fine-grained relations. A reason for the smaller increase could be due to the fact that some of the models the authors experiment with have large context windows (e.g., Gemini-2.5-flash has a 1 million token context window), but each article takes up on average between 15,000-30,000 tokens, so the model is still able to perform relatively well when provided with the total article's text.

\subsubsection{Edge Limitation}
One solution to the ``Overgeneration of Implicit or Incorrect Edges'' problem is to limit the number of edges the LLM can output, forcing it to focus on the surrounding textual context rather than relying on structure cues or token similarity. Edge limitation was implemented by adding an instruction to the prompt that the resulting derivation graphs should be returned with a maximum of two edges per node. Analysis on the distribution of the number of outgoing edges per node in the MDGD and found that the maximum number of outgoing edges per node in any article was $4$, and the average maximum was $1.65$. This prompted the authors to limit the number of edges to $2$, since the benefit of the LLMs not marking excessive edges outweighed the limited number of third- and fourth-outgoing edges the LLMs would miss under this rule. With edge limitation, the authors observe that precision, recall and $F_1$ score all slightly increase. This result is logical, as with edge limitations, LLM models make fewer predictions, resulting in more precise correct predictions. An increase in the $F_1$ score indicates that this restriction forces the model to focus on the most important edges when conducting mathematical reasoning.

\subsubsection{Post-Processing}
Another potential fix for the ``Overgeneration of Implicit or Incorrect Edges'' is a simple post-processing step to remove these redundant edges. For example, if the prediction identifies Eq(1) -> Eq(2), Eq(2)->Eq(3), and Eq(1)->Eq(3), the post-processing steps discards the last edge since Eq(3) is already in the reach of the derivation graph and is handled by the first two. The authors ran this post-processing step on two of the better performing methods, Brute Force + Gemini-2.5-flash, and Zero-Shot - GPT-5. From the performance metrics utilized, the authors found that the post-processing step mainly reduces recall, as it effectively extracts fewer edges, with a slight increase in precision and virtually no change with the and $F_1$ score.

\subsubsection{Prompt Engineering}
A significant aspect of getting LLMs to output the desired result is prompt engineering. Through prompt engineering, which the authors tested on the Gemini and GPT models, they devised multiple prompts that focused on different aspects of the derivation graph process. The authors focused on different prompts that emphasized minimal instruction, mathematical dependencies, or the hierarchical shape of derivation graphs (more details on prompts can be found in Appendix~\ref{apx:llm}). 
The hierarchical prompt achieved the highest conditional performance on the subset of responses that were correctly formatted and parsable, but had a very low success rate, with only $2$ of $107$ responses being correctly formatted. By contrast, the original prompt produced more parsable outputs ($92/107$) while maintaining relatively the same performance as the hierarchical prompt, thereby offering the best practical performance and overall utility. Given that an important part of LLM prompting involves responses being formatted accurately, the authors observe that the original prompt, which was direct and succinct, yields the best performance. The authors also experimented with structured output formatting such as JSON and found that there was not a statistically significant difference in the performance metrics compared to the requested formatting in the chosen prompt (difference of less than 1\% across precision, recall, and F1 score) and that the number of responses that were returned with a correctly parsable response was roughly the same ($\pm$ 1 article).

\subsubsection{Few-shot Experimentation}
To test whether providing contextual examples could improve derivation graph extraction for LLM models, the authors also conducted few-shot experimentation on the Gemini-2.5-flash and GPT-5 models. For these experiments, the authors used a \textbf{2-shot} setup: each prompt contained two problem-solution pairs, which were selected from the MDGD dataset. The problem-solution pairs were chosen to be representative of the task rather than random, prioritizing examples that covered (1) a diversity of derivation relation patterns, and (2) a diversity in the domain of the article's content. The prompt construction for the few-shot experimentation started with a short introduction, followed by the two examples (examples included the whole article text, followed by the hand-annotated derivation graph for the corresponding text in the same output format being requested), and ending with the text for the article that the experiment is being tested on. The authors maintained the same output format and prompt wording used in the initial zero-shot experimentation to ensure comparability.

\subsubsection{Potential Fixes Conclusion}
While these fixes offer some solutions to the LLM issues, they do not suffice. Other strategies, like combining fixes or developing a task-specific LLM, should be explored. From these results, the authors can conclude that constraining generation behavior, refining prompts, and integrating pattern-matching methods with models results in a slight improvement with derivation graph extraction. These strategies directly address limitations of current LLM-based approaches by enhancing completeness and precision. Future work should systematically evaluate these strategies and their trade-offs.

%% file: arxiv_preprint_2/sections/conclusion.tex
\section{Conclusion}
This paper introduces a novel task of summarizing mathematical literature by reconstructing dependency graphs that capture the derivation relationships between mathematical equations. These mathematical \textit{derivation graphs} summarize the logical flow of equations in STEM articles through a graphical representation. We have authored the Mathematical Derivation Graphs Dataset (MDGD), which comprises extracted derivation graphs from published arXiv manuscripts, and used it to evaluate the performance of various language processing algorithms. We evaluate analytical baseline models and LLMs on the relation extraction task of derivation graphs. Initial testing showed that these models reached $F_1$ scores of $\sim45\%-52\%$. After diagnosing failure points and applying fixes such as few-shot learning, and using a combination of analytical and LLM models together, the performance shows slight improvements with LLM performance. These results mark a promising first step in this challenging area of research. They also underscore the need for further improvements to advance the fields of mathematical relation extraction and mathematical language processing.

%% file: arxiv_preprint_2/sections/limitations.tex
\section{Limitations, Risks, and Ethical Considerations}

While our derivation graph formulation  advances the automation of mathematical article understanding, several limitations remain: (1) \textbf{Textual Context Window:} Analytic and Naive Bayes methods consider only local sentence windows (immediately before/after an equation), potentially overlooking long‑range dependencies. LLM prompts likewise have input‑length constraints that can truncate context. (2) \textbf{LLM Generality vs. Math Specificity:} Off‑the‑shelf LLMs were neither fine‑tuned nor designed specifically for relation extraction or symbolic reasoning, limiting their ability to handle nuanced mathematical language. Our prompt engineering partially mitigates this, but falls short of model‑level improvements. (3) \textbf{Future Work Dependencies:} Additional fixes, such as fine‑tuning, require a significant investment to generate a dataset large enough for the fine-tuning to impact the outputs significantly. Even with fine-tuning, improvement is not guaranteed. Additionally, this goes against the goal of the research since it would not test the general mathematical reasoning capabilities of LLMs. Fine-tuning was not implemented due to resource limitations. While our methods are intended as foundational research, risks exist in that LLM-based extractions could be misused. We therefore only use LLM outputs in the context of derivation graphs.

%% file: arxiv_preprint_2/appendix/dataset.tex
\newpage
\section{Dataset Details} \label{apx:data}

\begin{figure*}[htbp]
  \centering
  \begin{lstlisting}[language=json]
{   "Article ID": "1409.0466",
    "Equation ID": ["S3.E1",
                    "S3.E2",
                    "S3.E3",
                    "S3.E4",
                    "S3.E5",
                    "S5.E6",
                    "S5.E7" ],
    "Adjacency List": { "S3.E1": [ "S3.E3",
                                    "S3.E5" ],
                        "S3.E2": [  "S3.E3" ],
                        "S3.E3": [  null    ],
                        "S3.E4": [  "S3.E5" ],
                        "S3.E5": [  null    ],
                        "S5.E6": [  null    ],
                        "S5.E7": [  null    ] },
    "Equation Number": {"S3.E1": "1",
                        "S3.E2": "2",
                        "S3.E3": "3",
                        "S3.E4": "4",
                        "S3.E5": "5",
                        "S5.E6": "6",
                        "S5.E7": "7" } }
  \end{lstlisting}
  \caption{Data set entry example of article 1409.0466 \citep{comeron2014evidence}. Entries include the article ID, a list of equation IDs, the derivation graph adjacency list, the equation number, and the most important equation.}
  \label{fig:json_example}
\end{figure*}

Our dataset comprises of $107$ articles with around $2000$ equations. This dataset contains the hand-labeled derivation graphs for each article. Given the diversity in the dataset, further research could be conducted into how specific article structures for each category could be analyzed and utilized to improve the algorithms. An adjacency list of its derivation graph was stored in a singular JSON file, along with auxiliary information for each article. Examples of dataset entries can be seen in Fig.~\ref{fig:json_example}.

Each article’s metadata and derivation graph are serialized as a single JSON object with the following four fields:
\begin{itemize}
  \item \textbf{Article ID:} The arXiv identifier, e.g.\ \texttt{``1409.0466''}.
  \item \textbf{Equation ID:} An ordered list of all extracted equation IDs.
  \item \textbf{Adjacency List:} A mapping from each source equation ID to a list of target IDs it directly derives. A null entry (or empty array) denotes no outgoing edges.
  \item \textbf{Equation Number:} A human-readable index for each equation, corresponding to its appearance order in the article.
\end{itemize}

Due to the large data set size, the authors labeled the articles individually. To ensure validity, they checked each other's labeling both after the dataset was created and when performance numbers were unusual. When dealing with human annotators, it is inevitable that some subjectivity will be present. During annotation, the authors had multiple discussions about each article to ensure consistency with the annotation goals, which were to minimize the graph's complexity while maintaining the semantic structure of the underlying manuscript.

\begin{figure*} [htbp]
    \centering
    \fbox{\adjustbox{padding=5pt}{\includegraphics[width=0.9\linewidth]{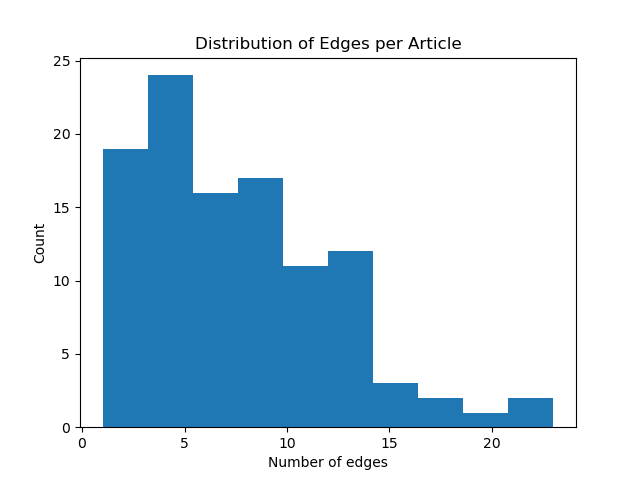}}}
    \caption{Distribution of Edges per Article in the MGDG}
    \label{fig:mdgd_hist_edge}
\end{figure*}

A histogram of the edges per article in the dataset can be found in Figure~\ref{fig:mdgd_hist_edge}, and specific numbers can be found in Table~\ref{tab:mdgd_dist}.

\begin{table*}[htbp]
  \caption{MDGD distribution}
  \label{tab:mdgd_dist}
  \centering
  \begin{tabular}{p{3cm}|cccc}
    \toprule
    \textbf{} & \textbf{Mean} & \textbf{Median} & \textbf{Min} &\textbf{Max} \\
    \midrule
    Equations per article  & $10.88$ & $10$ & $4$ & $21$ \\
    Edges per article & $7.47$ & $7$ & $1$ & $23$ \\
    Longest path (edges) & $3.06$ & $3$ & $1$ & $9$ \\
    \bottomrule
\end{tabular}
\end{table*}

%% file: arxiv_preprint_2/appendix/algos.tex
\section{Derivation Graph Algorithms} \label{apx:dga}

This appendix details the primary algorithms we evaluated for constructing derivation graphs of scientific articles. For each method, we describe its core intuition, computational complexity, key hyperparameters, and typical strengths and weaknesses observed in the experiments.

\subsection{Brute Force Algorithm}
This method exhaustively scans the article text for overt mentions of one equation within the context of another. It assumes that any explicit reference such as, ``From Eq.~(2) we derive Eq.~(5)'', signals a true derivation edge. An assumption in the brute-force algorithm is that equations that appear later in the paper will always be the parent of equations that appear later in the paper. For example, because Eq.~(2) is presented before Eq.~(5) chronologically, if there were to be a relationship between them, it would most likely be Eq.~(2) $\rightarrow$ Eq.~(5). The authors also considered the case of sentence phrasing (for example, ``From Eq.~(2) we derive Eq.~(5)'' vs. ``Eq.~(2) can be obtained from Eq.~(5)'') In an initial iteration of the algorithm the authors had a set of words that they would check occurred between two equation links, which would flip the direction of the edge (attempting to account for the example you have correctly pointed out). However, this led to a substantial decrease in the performance metrics, and the authors felt that without significantly more pre-processing, it would be better to stick with the chronological assumption without exception, so that the algorithm would serve as a rough baseline.

The implementation can be found with Algorithm~\ref{alg:brute}. It runs in $\mathcal{O}(N^2 \times S)$ time, where $N$ is the number of equations and $S$ is the average sentence length. This algorithm uses the following procedure:
\begin{enumerate}
  \item Iterate over each pair $(v_2, v_1)$ where $v_2$ appears before $v_1$ in the text.
  \item If $v_2$ is mentioned in the sentence containing $v_1$, add directed edge $v_2 \to v_1$.
  \item Continue until all pairs are checked.
\end{enumerate}

An advantage of such a method is that it has high precision when explicit derivation mentions exist, but falls short with implicit derivations and is prone to noise if equation mentions are abundant. 

\begin{algorithm*}[htbp]
\caption{Brute Force Algorithm}
\label{alg:brute}
\begin{algorithmic}
    \For{each vertex $v_1$}
        \For{each vertex $v_2$ preceding $v_1$}
            \If{$v_2$ explicitly mentioned before $v_1$ (in sentence $s_1$)}
                \State $E \gets (v_2, v_1)$
                \State $Adj[v_1] \gets Adj[v_1] + \{E\}$
            \EndIf
        \EndFor
    \EndFor
\end{algorithmic}
\end{algorithm*}

\subsection{Token Similarity Algorithm}

The intuition behind the token similarity algorithm is that derivations often preserve significant overlap of variable tokens and operators. By comparing token sets of two equations, we are able to infer a derivation when their shared token ratio exceeds a threshold. This algorithm uses the following procedure:
\begin{enumerate}
  \item Extract unique token sets $T_1, T_2$ from each equation pair $(e_1,e_2)$ by tokenizing at the character level.
  \item Compute overlap ratios
    \[
      r_{1\to2} = \frac{|T_1 \cap T_2|}{|T_1|},\quad
      r_{2\to1} = \frac{|T_1 \cap T_2|}{|T_2|}.
    \]
  \item If both $r_{1\to2}, r_{2\to1} > \tau$, direct the edge from the equation with larger ratio to the other.
\end{enumerate}

The implementation can be found with Algorithm~\ref{alg:equation_similarity}. The hyper-parameter threshold $\tau$ controls sensitivity: lower values increase recall but may introduce spurious edges. Additional hyper-parameters include the minimum number of ratios that must meet the threshold for an edge to be considered, but deals with similar problems of increasing recall but falling precision. This implementation runs in $O(N^2 \times T)$ where $N$ is the number of equations and $T$ is the typical token count per equation. The current tokenization method is simple and fast, but loses structural and mathematical information which could be holding it back. Experimentation with different tokenization methods could be pursued in further research.

An advantage of using the token similarity algorithm is that is recovers implicit structural relations, but suffers when equations share generic tokens or token similarity ratios are difficult to differentiate due to similar construction.

\begin{algorithm*}[htbp]
    \caption{Token Similarity Algorithm}
    \label{alg:equation_similarity}
    \begin{algorithmic}
        \State $equations \gets get\_equations(article)$ \Comment{get all equations for the current article}
        \State $n \gets num\_equations(equations)$ \Comment{number of equations}
        \State $threshold \gets \tau$ \Comment{hyperparameter threshold}
        \For{each equation $e_1$ in equations}
            \For{equation $e_2$ in equations, $e_1 \ne e_2$}
            \State $set_1 \gets unique\_set(e_1)$ \Comment{get all unique tokens in $e_1$}
            \State $set_2 \gets unique\_set(e_2)$ \Comment{get all unique tokens in $e_2$}
            \State $intersection\_set \gets intersection(set_1, set_2)$ \Comment{tokens that are the same in both sets}
            \State $percentage\_1\_in\_2 \gets \frac{|intersection\_set|}{|set_1|}$
            \State $percentage\_2\_in\_1 \gets \frac{|intersection\_set|}{|set_2|}$
            \If{$percentage\_1\_in\_2 > threshold$ \textbf{and} $percentage\_2\_in\_1 > threshold$}
                \If{$percentage\_1\_in\_2 > percentage\_2\_in\_1$}
                \Comment{determine edge direction}
                    \State $Adj[e_1] \gets e_2$
                \Else
                    \State $Adj[e_2] \gets e_1$
                \EndIf
            \EndIf
            \EndFor
        \EndFor
    \end{algorithmic}
\end{algorithm*}

\subsection{Naive Bayes Algorithm}

The Naive Bayes algorithm leverages supervised learning methods by constructing a classifier which predicts whether a derivation exists between any ordered equation pair, based on specified features. Naive Bayes assumes that, given a label, each word in a feature is conditionally independent from the others, a simplification that makes computation tractable but can miss dependencies between words that might be meaningful. We utilize the following procedure:
\begin{itemize}
  \item \textbf{Training:}  
    Extract labeled pairs from a gold‐standard corpus, compute feature vectors, and train a multinomial Naive Bayes model.
  \item \textbf{Inference:}  
    For a new article, generate all equation pairs’ feature vectors and predict a label (+1 for $e_i \to e_j$, –1 for $e_j \to e_i$, or 0 for no edge). Add directed edges accordingly.
\end{itemize}

The implementation can be found in Algorithm~\ref{alg:naive_bayes}. The labels and features for each pair of equations $i$ and $j$ (where $i < j$), were generated using the following formulation:
\begin{align}
\begin{gathered}
    \text{label}_{i, j} = \begin{cases} +1 & ,\text{edge: } i \rightarrow j \\ \phantom{+}0 & ,\text{no edge} \\ -1 & ,\text{edge: } i \leftarrow j \end{cases}
\end{gathered}
\end{align}
\begin{align}
\begin{gathered}
    \text{feature}_{i, j} = \text{equation}_i + \\ \sum_{k = i + 1}^{j} \text{words\_between\_equations}(k-1, k) \\+ \text{equation}_j
\end{gathered}
\end{align}
The intuition here is that any useful context for determining whether an edge exists, and in which direction, should appear either in the MathML alttext of the two equations or in the text between them. This approach treats each pair of equations as a standalone decision. If an article contains $n$ equations, then the models generates $\frac{n(n-1)}{2}$ derivation relationships. For the implementation, the feature input for the model is a bag-of-words count vector, and the model is MultinomialNB which is trained on these count vectors. The feature vector construction uses the CountVectorizer for tokenization and not transformer-based tokenizers, though these could be a point of interest for future experiments. We use MultinomialNB due to the multiclass labels of $+1, -1, 0$ for the relations. We utilize a hyper-parameter which controls the size of the training set. An advantage of the implementation includes that it adapts to varied linguistic cues and improves with more labeled data. However, the Naive Bayes classifier requires substantial annotation effort with an increased dataset size and is very performance sensitive to feature design and training data.

\begin{algorithm*}[htbp]
    \caption{Naive Bayes Algorithm}
    \label{alg:naive_bayes}
    \begin{algorithmic}
        \State \textbf{Training:}
        \State $train\_labels \gets [\text{ }]$ \Comment{initialize training label array}
        \State $train\_features \gets [\text{ }]$ \Comment{initialize training feature array}
        \For{each article $a$ in the training set }
            \State $equations \gets get\_equations(a)$ \Comment{get all equations for the current article}
            \State $current\_labels \gets extract\_labels(equations)$ \Comment{get labels for current article}
            \State $current\_features \gets extract\_features(equations)$ \Comment{get features for current article}
            \State $train\_labels \gets train\_labels + current\_labels$
            \State $train\_features \gets train\_features + current\_features$
        \EndFor
        \State $classifier \gets train\_naive\_bayes(train\_labels, train\_features)$ \Comment{train the Naive Bayes model}
        \State 
        \State \textbf{Testing:}
        \State $equations \gets get\_equations(article)$ \Comment{get all equations for the article being tested}
        \State $n \gets num\_equations(equations)$ \Comment{number of equations}
        \State $current\_features \gets extract\_features(equations)$ \Comment{get features for current article}
        \State $predicted\_labels \gets classifier.fit(current\_features)$ \Comment{get predicted labels}
        \State $label\_index \gets 0$ \Comment{index for current label prediction array}
        \For{equation $e_1 \gets e_1, e_2, e_3, \dots, e_n$}
            \For{equation $e_2 \gets e_{i+1}, e_{i + 2}, e_{i + 3}, \dots, e_n$}
                \If{$predicted\_labels[label\_index] = 1$}
                    \State $Adj[e_1] \gets e_2$ \Comment{edge $e_1 \rightarrow e_2$}
                \ElsIf{$predicted\_labels[label\_index] = -1$}
                    \State $Adj[e_2] \gets e_1$ \Comment{edge $e_1 \leftarrow e_2$}
                \EndIf
                \State $label\_index \gets label\_index + 1$
            \EndFor
        \EndFor
    \end{algorithmic}
\end{algorithm*}

%% file: arxiv_preprint_2/appendix/large_language_models.tex
\subsection{Large Language Models} \label{apx:llm}
In addition to heuristic and supervised approaches, we explore prompting large language models (LLMs) to infer derivation relations directly. Below we describe three of the alternate LLM prompt variants that we tried when coming up with our final prompt, each targeting a different aspect of prompt engineering:
\begin{itemize}
    \item \textbf{Minimal Instruction:}
    serves as a baseline prompt that intentionally limits guidance to test how well an LLM can infer structure and intent from sparse instruction. This approach mirrors zero-shot prompting and helps evaluate the model's inherent understanding of the task, though it may lead to incomplete or inconsistently formatted responses. The prompt can be seen in Fig.~\ref{fig:min_prompt-box}.
\begin{figure}[ht]
\centering
\begin{tcolorbox}[enhanced,
  attach boxed title to top center={yshift=-3mm,yshifttext=-1mm},
  colback=blue!5!white,
  colframe=blue!75!black,
  title=Minimal Instruction LLM Prompt]
``From the following article and its extracted equations:$\backslash$nArticle text:$\backslash$n''+\{total\_article\_text\}+``$\backslash$nEquations: $\backslash$n''+\{equation\_list\}+``$\backslash$n Identify dependencies among equations, providing results in this format:$\backslash$n EquationNumber -> DerivedEquations;$\backslash$n EquationNumberWithoutDependencies ->;$\backslash$n''.
  
\end{tcolorbox}
\caption{Minimal instruction prompt template for extracting equation derivations from an article.}
\label{fig:min_prompt-box}
\end{figure}

    \item \textbf{Mathematical Dependencies:} introduces domain-specific terminology, such as ``derived from'' and ``independent equations'', to provide more explicit semantic cues. The goal is to align the model's output format more closely with graph construction requirements, thereby reducing ambiguity in node and edge interpretation. This often improves consistency and relevance of dependencies compared to minimal instruction, but still falters with inconsistently formatted responses. The prompt can be seen in Fig.~\ref{fig:math_prompt-box}.

\begin{figure}[ht]
\centering
\begin{tcolorbox}[enhanced,
  attach boxed title to top center={yshift=-3mm,yshifttext=-1mm},
  colback=blue!5!white,
  colframe=blue!75!black,
  title=Mathematical Dependency LLM Prompt]
``Consider the equations listed below from a mathematical article:$\backslash$n''+\{equation\_list\}+``$\backslash$n Determine the relationships between the equations. Specifically, identify which equations are derived from others, based solely on the context. Output the results as: SourceEquation -> DerivedEquations;$\backslash$n IndependentEquation ->;$\backslash$n''.
  
\end{tcolorbox}
\caption{Mathematical dependency prompt template for extracting equation derivations from an article.}
\label{fig:math_prompt-box}
\end{figure}

    \item \textbf{Hierarchical Structure:} guides the model to construct a graph-like structure of dependencies, simulating how derivations often build upon intermediate results in layered steps. This prompt is particularly useful in cases where derivations are multi-stage or nested. However, due to its verbosity, the model may over-generate or infer deeper hierarchies than actually exist. This also results in inconsistently formatted responses, and the correctly formatted responses typically deal with a high number of false positive edges. The prompt can be seen in Fig.~\ref{fig:hier_prompt-box}.

\begin{figure}[ht]
\centering
\begin{tcolorbox}[enhanced,
  attach boxed title to top center={yshift=-3mm,yshifttext=-1mm},
  colback=blue!5!white,
  colframe=blue!75!black,
  title=Hierarchical Structure LLM Prompt]
``The following article includes hierarchical derivations of mathematical equations:$\backslash$n''+\{total\_article\_text\}+``$\backslash$n I have listed the equations extracted below:$\backslash$n''+\{equation\_list\}+``$\backslash$n Construct a hierarchy showing which equations are derived from others. Format the output as:$\backslash$n Root Equation -> Derived Equation 1, Derived Equation 2;$\backslash$nDerived Equation 1 -> Sub-Derived Equation 1, Sub-Derived Equation 2;$\backslash$n...$\backslash$nIf no further derivations exist, use empty space (e.g., Derived Equation 2 -> ).$\backslash$n''.
  
\end{tcolorbox}
\caption{Hierarchical structure prompt template for extracting equation derivations from an article.}
\label{fig:hier_prompt-box}
\end{figure}

\end{itemize}

%% file: arxiv_preprint_2/appendix/further_analysis.tex
\subsection{Further Analysis}
An interesting question raised is about classifying the ``hardness'' of derivation graphs and the subsequent performance of methods on easier and harder tasks. The authors analyzed the dataset by classifying hardness as the length of the longest derivation path in the ground-truth derivation graphs. A short analysis is as in Table.~\ref{tab:further_analysis}, looking at the Brute Force + Gemini-2.5-flash output.

\begin{table*}[h]
  \caption{Brute Force + Gemini-2.5-flash Derivation Graph Hardness Analysis}
  \label{tab:further_analysis}
  \centering
  \begin{tabular}{p{3cm}|cccc}
    \toprule
    \textbf{Length of Longest Derivation Path} & \textbf{MDGD Count}& \textbf{Mean Precision} & \textbf{Mean Recall} & \textbf{Mean $F_1$}\\
    \midrule
    $2$ & $\mathbf{33}$ & $43.3\%$ & $75.5\%$ & $52.1\%$ \\
    $0-1$ & $20$ & $29.0\%$ & $47.2\%$ & $28.6\%$ \\
    $3$ &  $20$ & $59.6\%$ & $66.9\%$ & $61.6\%$ \\
    $5-6$ & $18$ & $55.6\%$ & $65.9\%$ & $59.6\%$ \\
    $4$ & $11$ & $\mathbf{60.6}\%$ & $\mathbf{76.2}\%$ & $\mathbf{64.9\%}$ \\
    $7-10$ & $5$ & $30.3\%$ & $25.9\%$ & $26.7\%$ \\
    \bottomrule
\end{tabular}
\end{table*}

%% file: references.bib
@article{Zhou_2018,
   title={Testing conformal gravity with the supermassive black hole in 1H0707-495},
   volume={98},
   ISSN={2470-0029},
   url={http://dx.doi.org/10.1103/PhysRevD.98.024007},
   DOI={10.1103/physrevd.98.024007},
   number={2},
   journal={Physical Review D},
   publisher={American Physical Society (APS)},
   author={Zhou, Menglei and Cao, Zheng and Abdikamalov, Askar and Ayzenberg, Dimitry and Bambi, Cosimo and Modesto, Leonardo and Nampalliwar, Sourabh},
   year={2018},
   month=jul }

@misc{lala2023paperqaretrievalaugmentedgenerativeagent,
title={PaperQA: Retrieval-Augmented Generative Agent for Scientific Research},
author={Jakub Lála and Odhran O'Donoghue and Aleksandar Shtedritski and Sam Cox and Samuel G. Rodriques and Andrew D. White},
year={2023},
eprint={2312.07559},
archivePrefix={arXiv},
primaryClass={cs.CL},
url={https://arxiv.org/abs/2312.07559},
}

@article{Wang_2025,
title={SciDaSynth: Interactive Structured Data Extraction From Scientific Literature With Large Language Model},
volume={21},
ISSN={1891-1803},
url={http://dx.doi.org/10.1002/cl2.70073},
DOI={10.1002/cl2.70073},
number={4},
journal={Campbell Systematic Reviews},
publisher={Wiley},
author={Wang, Xingbo and Huey, Samantha L. and Sheng, Rui and Mehta, Saurabh and Wang, Fei},
year={2025},
month=nov }

@misc{zhou2024refactorlearningextracttheorems,
title={REFACTOR: Learning to Extract Theorems from Proofs},
author={Jin Peng Zhou and Yuhuai Wu and Qiyang Li and Roger Grosse},
year={2024},
eprint={2402.17032},
archivePrefix={arXiv},
primaryClass={cs.AI},
url={https://arxiv.org/abs/2402.17032},
}

@misc{zhu2024keypointdrivenmathematicalreasoningdistillation,
title={Key-Point-Driven Mathematical Reasoning Distillation of Large Language Model},
author={Xunyu Zhu and Jian Li and Can Ma and Weiping Wang},
year={2024},
eprint={2407.10167},
archivePrefix={arXiv},
primaryClass={cs.CL},
url={https://arxiv.org/abs/2407.10167},
}

@misc{chen2023programthoughtspromptingdisentangling,
title={Program of Thoughts Prompting: Disentangling Computation from Reasoning for Numerical Reasoning Tasks},
author={Wenhu Chen and Xueguang Ma and Xinyi Wang and William W. Cohen},
year={2023},
eprint={2211.12588},
archivePrefix={arXiv},
primaryClass={cs.CL},
url={https://arxiv.org/abs/2211.12588},
}

@article{10.1007/s10791-017-9296-8,
author = {Kristianto, Giovanni Yoko and Topi\'{c}, Goran and Aizawa, Akiko},
title = {Utilizing dependency relationships between math expressions in math IR},
year = {2017},
issue_date = {Apr 2017},
publisher = {Kluwer Academic Publishers},
address = {USA},
volume = {20},
number = {2},
issn = {1386-4564},
url = {https://doi.org/10.1007/s10791-017-9296-8},
doi = {10.1007/s10791-017-9296-8},
journal = {Inf. Retr.},
month = apr,
pages = {132–167},
numpages = {36}
}

@INPROCEEDINGS{9671942,
  author={Wang, Zichao and Zhang, Mengxue and Baraniuk, Richard G. and Lan, Andrew S.},
  booktitle={2021 IEEE International Conference on Big Data (Big Data)}, 
  title={Scientific Formula Retrieval via Tree Embeddings}, 
  year={2021},
  volume={},
  number={},
  pages={1493-1503},
  doi={10.1109/BigData52589.2021.9671942}}

@misc{wang2023legoproverneuraltheoremproving,
      title={LEGO-Prover: Neural Theorem Proving with Growing Libraries}, 
      author={Haiming Wang and Huajian Xin and Chuanyang Zheng and Lin Li and Zhengying Liu and Qingxing Cao and Yinya Huang and Jing Xiong and Han Shi and Enze Xie and Jian Yin and Zhenguo Li and Heng Liao and Xiaodan Liang},
      year={2023},
      eprint={2310.00656},
      archivePrefix={arXiv},
      primaryClass={cs.AI},
      url={https://arxiv.org/abs/2310.00656}, 
}

@misc{w3cMathML3,
  title        = {Mathematical Markup Language (MathML) Version 3.0 2nd Edition},
  author       = {David Carlisle and Patrick Ion and Robert Miner and Nico Poppelier},
  year         = {2014},
  howpublished = {World Wide Web Consortium (W3C) Recommendation},
  url          = {https://www.w3.org/TR/MathML3/}
}

@article{sarawagi2008information,
  title={Information extraction},
  author={Sarawagi, Sunita and others},
  journal={Foundations and Trends{\textregistered} in Databases},
  volume={1},
  number={3},
  pages={261--377},
  year={2008},
  publisher={Now Publishers, Inc.}
}

@article{ji2021survey,
  title={A survey on knowledge graphs: Representation, acquisition, and applications},
  author={Ji, Shaoxiong and Pan, Shirui and Cambria, Erik and Marttinen, Pekka and Yu, Philip S},
  journal={IEEE transactions on neural networks and learning systems},
  volume={33},
  number={2},
  pages={494--514},
  year={2021},
  publisher={IEEE}
}

@article{yu2020relationship,
  title={A relationship extraction method for domain knowledge graph construction},
  author={Yu, Haoze and Li, Haisheng and Mao, Dianhui and Cai, Qiang},
  journal={World Wide Web},
  volume={23},
  number={2},
  pages={735--753},
  year={2020},
  publisher={Springer}
}

@inproceedings{Chen_2022, series={WWW ’22},
   title={KnowPrompt: Knowledge-aware Prompt-tuning with Synergistic Optimization for Relation Extraction},
   url={http://dx.doi.org/10.1145/3485447.3511998},
   DOI={10.1145/3485447.3511998},
   booktitle={Proceedings of the ACM Web Conference 2022},
   publisher={ACM},
   author={Chen, Xiang and Zhang, Ningyu and Xie, Xin and Deng, Shumin and Yao, Yunzhi and Tan, Chuanqi and Huang, Fei and Si, Luo and Chen, Huajun},
   year={2022},
   month=apr, pages={2778–2788},
   collection={WWW ’22} }

@misc{qu2020fewshotrelationextractionbayesian,
      title={Few-shot Relation Extraction via Bayesian Meta-learning on Relation Graphs}, 
      author={Meng Qu and Tianyu Gao and Louis-Pascal A. C. Xhonneux and Jian Tang},
      year={2020},
      eprint={2007.02387},
      archivePrefix={arXiv},
      primaryClass={cs.LG},
      url={https://arxiv.org/abs/2007.02387}, 
}

@misc{zieword,
      title={Learning to Reason Deductively: Math Word Problem Solving as Complex Relation Extraction}, 
      author={Zhanming Jie and Jierui Li and Wei Lu},
      year={2022},
      eprint={2203.10316},
      archivePrefix={arXiv},
      primaryClass={cs.CL},
      url={https://arxiv.org/abs/2203.10316}, 
}

@article{10.1007/s10489-024-05985-y,
author = {Dai, Qizhu and Li, Rongzhen and Xue, Zhongxuan and Li, Xue and Zhong, Jiang},
title = {Document-level relation extraction via commonsense knowledge enhanced graph representation learning: Document-level relation extraction via commonsense knowledge...},
year = {2024},
issue_date = {Jan 2025},
publisher = {Kluwer Academic Publishers},
address = {USA},
volume = {55},
number = {2},
issn = {0924-669X},
url = {https://doi.org/10.1007/s10489-024-05985-y},
doi = {10.1007/s10489-024-05985-y},
journal = {Applied Intelligence},
month = dec,
numpages = {13}
}

@inproceedings{wadhwa-etal-2023-revisiting,
    title = "Revisiting Relation Extraction in the era of Large Language Models",
    author = "Wadhwa, Somin  and
      Amir, Silvio  and
      Wallace, Byron",
    editor = "Rogers, Anna  and
      Boyd-Graber, Jordan  and
      Okazaki, Naoaki",
    booktitle = "Proceedings of the 61st Annual Meeting of the Association for Computational Linguistics (Volume 1: Long Papers)",
    month = jul,
    year = "2023",
    address = "Toronto, Canada",
    publisher = "Association for Computational Linguistics",
    url = "https://aclanthology.org/2023.acl-long.868/",
    doi = "10.18653/v1/2023.acl-long.868",
    pages = "15566--15589"
}

@misc{wan2023gptreincontextlearningrelation,
      title={GPT-RE: In-context Learning for Relation Extraction using Large Language Models}, 
      author={Zhen Wan and Fei Cheng and Zhuoyuan Mao and Qianying Liu and Haiyue Song and Jiwei Li and Sadao Kurohashi},
      year={2023},
      eprint={2305.02105},
      archivePrefix={arXiv},
      primaryClass={cs.CL},
      url={https://arxiv.org/abs/2305.02105}, 
}

@misc{zhong2021frustratinglyeasyapproachentity,
      title={A Frustratingly Easy Approach for Entity and Relation Extraction}, 
      author={Zexuan Zhong and Danqi Chen},
      year={2021},
      eprint={2010.12812},
      archivePrefix={arXiv},
      primaryClass={cs.CL},
      url={https://arxiv.org/abs/2010.12812}, 
}

@misc{zhan2019spanmodelopeninformation,
      title={Span Model for Open Information Extraction on Accurate Corpus}, 
      author={Junlang Zhan and Hai Zhao},
      year={2019},
      eprint={1901.10879},
      archivePrefix={arXiv},
      primaryClass={cs.CL},
      url={https://arxiv.org/abs/1901.10879}, 
}

@inproceedings{stanovsky-etal-2018-supervised,
    title = "Supervised Open Information Extraction",
    author = "Stanovsky, Gabriel  and
      Michael, Julian  and
      Zettlemoyer, Luke  and
      Dagan, Ido",
    editor = "Walker, Marilyn  and
      Ji, Heng  and
      Stent, Amanda",
    booktitle = "Proceedings of the 2018 Conference of the North {A}merican Chapter of the Association for Computational Linguistics: Human Language Technologies, Volume 1 (Long Papers)",
    month = jun,
    year = "2018",
    address = "New Orleans, Louisiana",
    publisher = "Association for Computational Linguistics",
    url = "https://aclanthology.org/N18-1081/",
    doi = "10.18653/v1/N18-1081",
    pages = "885--895"
}

@inproceedings{qin2017,
  author={Qin, Pengda and Xu, Weiran and Guo, Jun},
  booktitle={2017 International Joint Conference on Neural Networks (IJCNN)}, 
  title={Designing an adaptive attention mechanism for relation classification}, 
  year={2017},
  volume={},
  pages={4356-4362},
  doi={10.1109/IJCNN.2017.7966407}}

@inproceedings{jiang-etal-2016-relation,
    title = "Relation Extraction with Multi-instance Multi-label Convolutional Neural Networks",
    author = "Jiang, Xiaotian  and
      Wang, Quan  and
      Li, Peng  and
      Wang, Bin",
    editor = "Matsumoto, Yuji  and
      Prasad, Rashmi",
    booktitle = "Proceedings of {COLING} 2016, the 26th International Conference on Computational Linguistics: Technical Papers",
    month = dec,
    year = "2016",
    address = "Osaka, Japan",
    publisher = "The COLING 2016 Organizing Committee",
    url = "https://aclanthology.org/C16-1139/",
    pages = "1471--1480"
}

@inproceedings{riloff1993automatically,
  title={Automatically constructing a dictionary for information extraction tasks},
  author={Riloff, Ellen and others},
  booktitle={AAAI},
  volume={1},
  pages={2--1},
  year={1993}
}

@inproceedings{hobbs1993fastus,
  title={FASTUS: A finite-state processor for information extraction from real-world text},
  author={Hobbs, Jerry},
  booktitle={… JOINT CONFERENCE ON…},
  year={1993}
}

@article{10.1145/1409360.1409378,
author = {Etzioni, Oren and Banko, Michele and Soderland, Stephen and Weld, Daniel S.},
title = {Open information extraction from the web},
year = {2008},
issue_date = {December 2008},
publisher = {Association for Computing Machinery},
address = {New York, NY, USA},
volume = {51},
number = {12},
issn = {0001-0782},
url = {https://doi.org/10.1145/1409360.1409378},
doi = {10.1145/1409360.1409378},
journal = {Commun. ACM},
month = dec,
pages = {68–74},
numpages = {7}
}

@inproceedings{ciravegna2001adaptive,
author = {Ciravegna, Fabio},
title = {Adaptive information extraction from text by rule induction and generalisation},
year = {2001},
isbn = {1558608125},
publisher = {Morgan Kaufmann Publishers Inc.},
address = {San Francisco, CA, USA},
booktitle = {Proceedings of the 17th International Joint Conference on Artificial Intelligence - Volume 2},
pages = {1251–1256},
numpages = {6},
location = {Seattle, WA, USA},
series = {IJCAI'01}
}

@inproceedings{muslea1999extraction,
  title={Extraction patterns for information extraction tasks: A survey},
  author={Muslea, Ion and others},
  booktitle={The AAAI-99 workshop on machine learning for information extraction},
  volume={2},
  year={1999},
  organization={Orlando Florida}
}

@inproceedings{grishman-sundheim-1996-message,
    title = "{M}essage {U}nderstanding {C}onference- 6: A Brief History",
    author = "Grishman, Ralph  and
      Sundheim, Beth",
    booktitle = "{COLING} 1996 Volume 1: The 16th International Conference on Computational Linguistics",
    year = "1996",
    url = "https://aclanthology.org/C96-1079/"
}

@inproceedings{marsh-perzanowski-1998-muc,
    title = "{MUC}-7 Evaluation of {IE} Technology: Overview of Results",
    author = "Marsh, Elaine  and
      Perzanowski, Dennis",
    booktitle = "Seventh Message Understanding Conference ({MUC}-7): Proceedings of a Conference Held in Fairfax, Virginia, {A}pril 29 - May 1, 1998",
    year = "1998",
    url = "https://aclanthology.org/M98-1002/"
}

@proceedings{10.5555/1071958,
title = {MUC3 '91: Proceedings of the 3rd conference on Message understanding},
year = {1991},
isbn = {1558602364},
organization = {Association for Computational Linguistics},
address = {USA},
location = {San Diego, California}
}

@proceedings{10.5555/1072017,
title = {MUC5 '93: Proceedings of the 5th conference on Message understanding},
year = {1993},
isbn = {1558603360},
organization = {Association for Computational Linguistics},
address = {USA},
location = {Baltimore, Maryland}
}

@proceedings{10.5555/1072064, title = {MUC4 '92: Proceedings of the 4th conference on Message understanding}, year = {1992}, isbn = {1558602739}, organization = {Association for Computational Linguistics}, address = {USA}, location = {McLean, Virginia} }

@inproceedings{chinchor-1998-overview,
    title = "Overview of {MUC}-7",
    author = "Chinchor, Nancy A.",
    booktitle = "Seventh Message Understanding Conference ({MUC}-7): Proceedings of a Conference Held in Fairfax, Virginia, {A}pril 29 - May 1, 1998",
    year = "1998",
    url = "https://aclanthology.org/M98-1001/"
}

@proceedings{10.5555/1072399,
title = {MUC6 '95: Proceedings of the 6th conference on Message understanding},
year = {1995},
isbn = {1558604022},
organization = {Association for Computational Linguistics},
address = {USA},
location = {Columbia, Maryland}
}

@article{DETROJA2023200244,
title = {A survey on Relation Extraction},
journal = {Intelligent Systems with Applications},
volume = {19},
pages = {200244},
year = {2023},
issn = {2667-3053},
doi = {https://doi.org/10.1016/j.iswa.2023.200244},
url = {https://www.sciencedirect.com/science/article/pii/S2667305323000698},
author = {Kartik Detroja and C.K. Bhensdadia and Brijesh S. Bhatt}
}

@article{bretzel2009barnett,
  title={Barnett effect in thin magnetic films and nanostructures},
  author={Bretzel, Stefan and Bauer, Gerrit EW and Tserkovnyak, Yaroslav and Brataas, Arne},
  journal={Applied Physics Letters},
  volume={95},
  number={12},
  year={2009},
  publisher={AIP Publishing}
}

@article{zhu2017local,
  title={A local leaky-box model for the local stellar surface density--gas surface density--gas phase metallicity relation},
  author={Zhu, Guangtun Ben and Barrera-Ballesteros, Jorge K and Heckman, Timothy M and Zakamska, Nadia L and S{\'a}nchez, Sebastian F and Yan, Renbin and Brinkmann, Jonathan},
  journal={Monthly Notices of the Royal Astronomical Society},
  volume={468},
  number={4},
  pages={4494--4501},
  year={2017},
  publisher={Oxford University Press}
}

@misc{arxiv,
  author={{arXiv}},
  title={arXiv e-Print Archive},
  howpublished={\url{https://arxiv.org}},
  year={2024}
}

@misc{arXMLiv,
  author={Michael Kohlhase and Others},
  title={The arXMLiv Project},
  howpublished={\url{https://kwarc.info/projects/arXMLiv/}},
  year={2024}
}

@article{comeron2014evidence,
  title={Evidence for the concurrent growth of thick discs and central mass concentrations from S4G imaging},
  author={Comer{\'o}n, S{\'e}bastien and Elmegreen, Bruce G and Salo, Heikki and Laurikainen, Eija and Holwerda, Benne W and Knapen, Johan H},
  journal={Astronomy \& Astrophysics},
  volume={571},
  pages={A58},
  year={2014},
  publisher={EDP Sciences}
}

@inproceedings{mccallum1998comparison,
  title={A comparison of event models for naive bayes text classification},
  author={McCallum, Andrew and Nigam, Kamal and others},
  booktitle={AAAI-98 workshop on learning for text categorization},
  volume={752},
  pages={41--48},
  year={1998},
  organization={Madison, WI}
}

@misc{math_derivation_graphs,
  author={Prasad, Vishesh and Kim, Brian and Kani, Nickvash},
  title={Mathematical Derivation Graphs Dataset (MDGD)},
  howpublished={\url{https://github.com/visheshprasad/Mathematical_Derivation_Graphs_Dataset_MDGD}},
  year={2025}
}

@misc{touvron2023llama2openfoundation,
      title={Llama 2: Open Foundation and Fine-Tuned Chat Models}, 
      author={Hugo Touvron and Louis Martin and Kevin Stone and Peter Albert and Amjad Almahairi and Yasmine Babaei and Nikolay Bashlykov and Soumya Batra and Prajjwal Bhargava and Shruti Bhosale and Dan Bikel and Lukas Blecher and Cristian Canton Ferrer and Moya Chen and Guillem Cucurull and David Esiobu and Jude Fernandes and Jeremy Fu and Wenyin Fu and Brian Fuller and Cynthia Gao and Vedanuj Goswami and Naman Goyal and Anthony Hartshorn and Saghar Hosseini and Rui Hou and Hakan Inan and Marcin Kardas and Viktor Kerkez and Madian Khabsa and Isabel Kloumann and Artem Korenev and Punit Singh Koura and Marie-Anne Lachaux and Thibaut Lavril and Jenya Lee and Diana Liskovich and Yinghai Lu and Yuning Mao and Xavier Martinet and Todor Mihaylov and Pushkar Mishra and Igor Molybog and Yixin Nie and Andrew Poulton and Jeremy Reizenstein and Rashi Rungta and Kalyan Saladi and Alan Schelten and Ruan Silva and Eric Michael Smith and Ranjan Subramanian and Xiaoqing Ellen Tan and Binh Tang and Ross Taylor and Adina Williams and Jian Xiang Kuan and Puxin Xu and Zheng Yan and Iliyan Zarov and Yuchen Zhang and Angela Fan and Melanie Kambadur and Sharan Narang and Aurelien Rodriguez and Robert Stojnic and Sergey Edunov and Thomas Scialom},
      year={2023},
      eprint={2307.09288},
      archivePrefix={arXiv},
      primaryClass={cs.CL},
      url={https://arxiv.org/abs/2307.09288}, 
}

@misc{mistral2024,
  title        = {Mistral NeMO},
  author       = {{Mistral AI Team}},
  year         = {2024},
  url          = {https://mistral.ai/news/mistral-nemo/}
}

@misc{hui2024qwen25codertechnicalreport,
      title={Qwen2.5-Coder Technical Report}, 
      author={Binyuan Hui and Jian Yang and Zeyu Cui and Jiaxi Yang and Dayiheng Liu and Lei Zhang and Tianyu Liu and Jiajun Zhang and Bowen Yu and Keming Lu and Kai Dang and Yang Fan and Yichang Zhang and An Yang and Rui Men and Fei Huang and Bo Zheng and Yibo Miao and Shanghaoran Quan and Yunlong Feng and Xingzhang Ren and Xuancheng Ren and Jingren Zhou and Junyang Lin},
      year={2024},
      eprint={2409.12186},
      archivePrefix={arXiv},
	Volume = {6}
}

@misc{fernández2010smallamplitudeapproximationdifferentialequation,
      title={On the small--amplitude approximation to the differential equation $\ddot{x}+(1+\dot{x}^{2})x=0$}, 
      author={Francisco M. Fernández},
      year={2010},
      eprint={0907.3505},
      archivePrefix={arXiv},
      primaryClass={math-ph},
      url={https://arxiv.org/abs/0907.3505}, 
}

@misc{haider2024phi3safetyposttrainingaligning,
      title={Phi-3 Safety Post-Training: Aligning Language Models with a "Break-Fix" Cycle}, 
      author={Emman Haider and Daniel Perez-Becker and Thomas Portet and Piyush Madan and Amit Garg and Atabak Ashfaq and David Majercak and Wen Wen and Dongwoo Kim and Ziyi Yang and Jianwen Zhang and Hiteshi Sharma and Blake Bullwinkel and Martin Pouliot and Amanda Minnich and Shiven Chawla and Solianna Herrera and Shahed Warreth and Maggie Engler and Gary Lopez and Nina Chikanov and Raja Sekhar Rao Dheekonda and Bolor-Erdene Jagdagdorj and Roman Lutz and Richard Lundeen and Tori Westerhoff and Pete Bryan and Christian Seifert and Ram Shankar Siva Kumar and Andrew Berkley and Alex Kessler},
      year={2024},
      eprint={2407.13833},
      archivePrefix={arXiv},
      primaryClass={cs.CL},
      url={https://arxiv.org/abs/2407.13833}, 
}

@misc{openai2025gpt5,
author = {OpenAI},
title = {GPT-5},
year = {2025},
note = {Large language model. Available at \url{https://platform.openai.com/}},
howpublished = {\url{https://platform.openai.com/}}
}
